# Importance Sampling in Bayesian Networks: An Influence-Based Approximation Strategy for Importance Functions

Changhe Yuan and Marek J. Druzdzel
Decision Systems Laboratory
Intelligent Systems Program and School of Information Sciences
University of Pittsburgh
Pittsburgh, PA 15260
{cyuan,marek}@sis.pitt.edu

## Abstract

One of the main problems of importance sampling in Bayesian networks is representation of the importance function, which should ideally be as close as possible to the posterior joint distribution. Typically, we represent an importance function as a factorization, i.e., product of conditional probability tables (CPTs).[1] Given diagnostic evidence, we do not have explicit forms for the CPTs in the networks. We first derive the exact form for the CPTs of the optimal importance function. Since the calculation is hard, we usually only use their approximations. We review several popular approximation strategies and point out their limitations. Based on an analysis of the influence of evidence, we propose a method for approximating the exact form of importance function by explicitly modeling the most important additional dependence relations introduced by evidence. Our experimental results show that the new approximation strategy offers an immediate improvement in the quality of the importance function.

## 1 Introduction

Importance sampling-based algorithms have become an important family of approximate inference methods for Bayesian networks. One of the main problems of importance sampling in Bayesian networks is representation of the importance function. It is well known that the quality of importance function is critical to importance sampling algorithms: The closer the importance function to the actual posterior distribution, the better its performance. Typically, we represent an importance function as a factorization, i.e., a product of *conditional probability tables* (CPTs). Given diagnostic evidence, additional dependence relations will be introduced among the variables. Consequently we do not have explicit forms for the CPTs in the networks. We first derive the exact form for the CPTs of the optimal importance function. Since calculating them is practically equivalent to exact inference in the networks, we usually only use their approximations. We review several existing approximation strategies for the conditional forms and point out their limitations. After a simple analysis of the influence of evidence in Bayesian networks, we propose an approximation strategy that modifies the network structure in order to accommodate the most important additional dependence relations introduced by the evidence. Our experimental results show that the new strategy offers an immediate improvement in the quality of the importance function.

## 2 Importance Sampling

We start with the theoretical roots of importance sampling. We use capital letters for variables and lowercase letters for their states. Bold letters denote sets of variables or states. Let $f(\mathbf{X})$ be a function of $n$ variables $\mathbf{X} = \{X_1, ..., X_n\}$ over the domain $\Omega \subset R^n$. Consider the problem of estimating the multiple integral

$$V = \int_\Omega f(\mathbf{x})d\mathbf{x} \ . \tag{1}$$

We assume that the domain of integration of $f(\mathbf{X})$ is bounded, i.e., that $V$ exists. Importance sampling approaches this problem by estimating

$$V = \int_\Omega \frac{f(\mathbf{x})}{g(\mathbf{x})} g(\mathbf{x})d\mathbf{x} \ , \tag{2}$$

where $g(\mathbf{X})$, which is called the *importance function*, is a probability density function such that $g(\mathbf{X}) > 0$ across the entire domain $\Omega$. One practical requirement of $g(\mathbf{X})$ is that it should be easy to sample from. In order to estimate the integral, we generate samples $\mathbf{x}_1, \mathbf{x}_2, ..., \mathbf{x}_N$ from $g(\mathbf{X})$ and use the generated values in the sample-mean formula

$$\hat{V} = \frac{1}{N} \sum_{i=1}^{N} \frac{f(\mathbf{x}_i)}{g(\mathbf{x}_i)} . \quad (3)$$

---

[1] Here we only consider discrete case; However, some of our results are also applicable for continuous case.

The estimator almost surely converges to $V$ under certain weak assumptions [4].

The performance of the estimator in Equation 3 can be measured by its variance

$$var_g[\frac{f(\mathbf{X})}{g(\mathbf{X})}] = \int_\Omega \frac{f^2(\mathbf{x})}{g(\mathbf{x})} d\mathbf{x} - V^2. \quad (4)$$

Rubinstein [16] shows that if $f(\mathbf{X}) > 0$, the optimal importance function is

$$g(\mathbf{X}) = \frac{f(\mathbf{X})}{V} . \quad (5)$$

In this case, the variance of the estimator is zero. However, the concept of the optimal importance function is of rather theoretical significance because finding $V$ is equivalent to finding the posterior distribution, which is the problem that we are facing. Nevertheless, it suggests that if we find instead a function that is close enough to the optimal importance function, we can still expect good convergence rates.

## 3 A General Form for Importance Functions in Bayesian Networks

Importance sampling can be easily adapted to solve a variety of inference problems in Bayesian networks, especially finding posterior marginals for unobserved variables. To make importance sampling in Bayesian networks practical, we typically need a sequence of CPTs, which jointly specify an importance function. More formally, suppose $P(\mathbf{X})$ models the joint probability distribution over a set of variables $\{X_1, X_2, ..., X_n\}$. By the chain rule, we can factorize it as follows.

$$P(\mathbf{X}) = P(X_1) \prod_{i=2}^{n} P(X_i | X_1, ..., X_{i-1}) . \quad (6)$$

To draw a sample for $\mathbf{X}$, we draw samples from each of the CPT $P(X_i | X_1, ..., X_{i-1})$ sequentially. We can easily get the CPTs in Bayesian networks with no evidence, because if $X_1, X_2, ..., X_n$ are in the topological order of the network, we can simplify the above equation to

$$P(\mathbf{X}) = \prod_{i=1}^{n} P(X_i | PA(X_i)) , \quad (7)$$

where $P(X_i | PA(X_i))$ are explicitly modeled in Bayesian networks. The above simplification reflects so called *Markov condition* [14], which says that a node is independent of its non-descendants given only its parents. Importance sampling under such circumstances is easy to implement. However, when diagnostic evidence exists, it can dramatically change the dependence relations among the variables. Suppose that in addition to the unobserved variables $\mathbf{X}$, we also have an evidence set $\mathbf{E} = \{E_1, ..., E_m\}$. We know that the posterior distribution of the network can still be factorized using the chain rule.

$$P(\mathbf{X}|\mathbf{E}) = P(X_1|\mathbf{E}) \prod_{i=2}^{n} P(X_i | X_1, ..., X_{i-1}, \mathbf{E}) . \quad (8)$$

However, the simplification made for Equation 7 can no longer be made here, because we cannot just throw away the variables in $\{X_1, ..., X_{i-1}\} \setminus PA(X_i)$, on some of which $X_i$ may depend given the evidence. Before we analyze how to simplify $P(X_i | X_1, ..., X_{i-1})$, we first introduce the following definition.

**Definition 1** *Given a Bayesian network with unobserved variables $\mathbf{X} = \{X_1, ..., X_n\}$ and evidence set $\mathbf{E}$. For an ordering of $X_1, ..., X_n$, the* relevant factor *of $X_i$, denoted as $RF(X_i)$, is the set of variables that appear before $X_i$ in the ordering and are d-connected [14] to $X_i$ conditional on the parents of $X_i$, the evidence $\mathbf{E}$, and the other variables in $RF(X_i)$.*

Intuitively, $RF(X_i)$ includes the additional variables that $X_i$ needs to condition on in $\{X_1, ..., X_{i-1}\}$. Note that $RF(X_i)$ is specific to a particular ordering of the variables; It may contain different variables for different orderings. Given the definition, Equation 8 can now be simplified to

$$P(\mathbf{X}|\mathbf{E}) = \prod_{i=1}^{n} P(X_i | PA(X_i), \mathbf{E}, RF(X_i)) . \quad (9)$$

However, we now have no explicit forms of the CPTs in the network any more. If we want to compute and store the CPTs, we need to break the constraint of the original network structure and accommodate the additional dependence among the variables. One solution is to construct a new network in which each node $X_i$ has arcs coming from the variables in both $PA(X_i)$ and $RF(X_i)$. We call such network *factorizable*.

**Definition 2** *Given a Bayesian network with unobserved variables $\mathbf{X} = \{X_1, ..., X_n\}$ and evidence set $\mathbf{E}$. A* factorizable *network is a new network that represents the same distribution as the original network and whose posterior distribution $P(\mathbf{X}|\mathbf{E})$ can be fully factorized to a product of CPTs, one for each unobserved variable $X_i$ in $\mathbf{X}$ conditional on its parents $PA(X_i)$.*

Factorizable structure is not unique. One algorithm to construct a factorizable structure for a Bayesian network is described below.

**Algorithm 1** *Building a Factorizable Structure.*

**Input**: *Bayesian network B, a set of evidence variables* **E**, *and a set of unobserved variables* **X**;

**Output**: *A factorizable structure.*

1. Order the nodes in **X** in the reverse of their topological order.

2. Mark the nodes that are ancestors of evidence nodes in **E**.

3. For the ordering in Step 1, check each unobserved node if it is marked. If so, add an arc between each pair of its parents given that no arc exists between them, such that the orientation of the arc is from the node appearing later in the order to the earlier one.

4. When adding an arc between two nodes, we expand the CPT of the child by duplicating the entries for different states of the parent.

It is straightforward to prove the following theorem.

**Theorem 1** *Applying Algorithm 1 to a Bayesian network with evidence* **E** *and unobserved variables* **X** $= \{X_1, ..., X_n\}$ *yields a factorizable structure.*

Algorithm 1 is similar to the *graph reduction* method proposed in [11, 18]. Indeed, our procedure will introduce the same additional arcs as the graph reduction method if two methods use the same ordering. However, the difference is that graph reduction absorbs the evidence while reducing the graph, which, in the end, results in a new network without evidence variables. In our procedure, we separate graph reduction and evidence absorption. We first create a new network that still represent the same distribution as the original one. Given the new structure, we can factorize the full joint posterior distribution using chain rule and absorb evidence into each CPT separately. Ortiz [12] presents a similar idea for constructing an optimal importance function, in which he suggests first triangulating the Bayesian network and making it chordal, and then constructing the new structure from the chordal graph. Our approach avoids his intermediate step.

In Step 3 of Algorithm 1, we add arcs between all the parents of a node. Hence, the last parent will get arcs coming from all the other parents. If the CPT size of the last parent is initially large, or if there are many parents, the new CPT for the last parent may blow up. Even not, the new structure will make importance sampling inefficient. To remedy the problem, we propose several heuristics for preprocessing the ordering of the parents. All the heuristics are subject to the partial constraints of the original network, which include arcs or directed paths that already exist among parents. The first method is to order the parents in the descending order of the number of their own parents. By doing so, we are trying to make the last parent have as less incoming arcs as possible, which can reduce the size of the CPTs. The second method is to order the parents in a descending order of the size of their CPTs. Since our purpose is to minimize the size of CPTs, the second heuristic is more effective, which we will use in all the experiments of this paper.

## 4 Previous Approximation Strategies for Importance Functions

From Equation 9, we can see that to build a factorizable structure, we need to add many arcs to the new structure when we have diagnostic evidence. These extra arcs make a network more complex and make the calculation of the CPTs much more difficult. Although we propose some heuristics for minimizing the size of CPTs, these can still grow too large. In fact, this process is practically equivalent to exact inference in the network. Therefore, we usually only use approximations of the full forms. Here we will review several approximation strategies used by the existing importance sampling algorithms for Bayesian networks. We will use a running example to illustrate these strategies.

**Example:** A simple Bayesian network with three binary variables in Figure 1 parameterized as follows:

| $a$ | 0.2 |
|---|---|
| $\neg a$ | 0.8 |

| $b$ | 0.7 |
|---|---|
| $\neg b$ | 0.3 |

| $P(C\|A,B)$ | $a$ | | $\neg a$ | |
|---|---|---|---|---|
| | $b$ | $\neg b$ | $b$ | $\neg b$ |
| c | 0.99 | 0.01 | 0.1 | 0.9 |
| $\neg c$ | 0.01 | 0.99 | 0.9 | 0.1 |

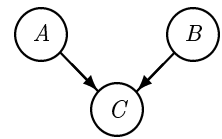

Figure 1: A simple Bayesian network.

Variable $C$ is observed in state $\neg c$. We can easily calculate the posterior joint distribution over $A$ and $B$.

| $P(A,B|\neg c)$ | $a$ | $\neg a$ |
|---|---|---|
| $b$ | 0.0024 | 0.8560 |
| $\neg b$ | 0.1009 | 0.0408 |

□

**Original CPT-based Importance Function**: Probabilistic logic sampling [6] assumes that the importance function has the same CPTs for all the variables. Likelihood weighting [3, 19] goes a step further and assumes that the importance function has the following form.

$$P(\mathbf{X}|\mathbf{E}) = \prod_{i=1}^{n} P(X_i|PA(X_i)\backslash\mathbf{E}, \mathbf{E} \cap PA(X_i)) \quad (10)$$

It uses the same CPTs as the original distribution for nodes with no evidence parents. Otherwise, it shrinks CPTs for those nodes with evidence parents. Obviously, this approximation only takes into account the primitive influence of the evidence, the influence of the evidence nodes on the CPTs of their children.

**Example:** Original CPT-based Importance Function for the running example.

| $P(A,B)$ | $a$ | $\neg a$ |
|---|---|---|
| $b$ | 0.14 | 0.56 |
| $\neg b$ | 0.06 | 0.24 |

□

**ICPT-based Importance Function**: Several algorithms notice the limitations of the importance function used by likelihood weighting and propose a different form of importance function. They still assume the same structure for the importance function as the original Bayesian network, but they realize that the evidence has influence on the CPTs of all the nodes and propose the following form of importance function.

$$P(\mathbf{X}|\mathbf{E}) = \prod_{i=1}^{n} P(X_i|PA(X_i)\backslash\mathbf{E}, \mathbf{E}). \quad (11)$$

Each $P(X_i|PA(X_i)\backslash\mathbf{E}, \mathbf{E})$ is called an *importance CPT* (ICPT), a concept first proposed in [1]. However, there are actually many algorithms that use the above importance function, in spite of the fact that they differ in the methods of estimating the actual tables. Several dynamic importance sampling algorithms, including AIS-BN [1], self-importance sampling [19], and adaptive importance sampling [13], use different learning methods to learn the ICPTs. Yuan and Druzdzel [20] derive the formula of calculating the ICPTs using belief propagation messages [10, 14]. The importance function in Equation 11 offers a big improvement over the representation of Equation 10. However, this representation still only takes into account partial influence of the evidence. In case the evidence dramatically changes dependence relations among the variables, this approximation will be suboptimal as well.

**Example:** ICPT-based Importance Function for the running example.

| $P(A,B)$ | $a$ | $\neg a$ |
|---|---|---|
| $b$ | 0.0886 | 0.7697 |
| $\neg b$ | 0.0146 | 0.1270 |

□

A dynamic importance sampling algorithm using the above form may learn a different importance function, depending on what distance measure it tries to minimize. For the running example, we need two parameters to parameterize the importance function. If we use K-L divergence as the distance measure, we get the same solutions as above. If we minimize the variance of the importance sampling estimator, the learned importance function has the following joint probability distribution.

**Example:** Importance Function learned by minimizing the variance of the importance sampling estimator for the running example.

| $P(A,B)$ | $a$ | $\neg a$ |
|---|---|---|
| $b$ | 0.1531 | 0.6487 |
| $\neg b$ | 0.0379 | 0.1604 |

□

We can see that we cannot achieve the optimal importance function by learning. The reason is that the actual posterior distribution typically needs more parameters to parameterize than the importance function. For the running example, the actual posterior distribution needs three parameters. Obviously, it is in general impossible to perfectly fit a three-parameter distribution with a two-parameter distribution.

**Variable Elimination-based Importance Function**: Hernandez et al. [8] propose to use the variable elimination [21] algorithm to eliminate the variables one by one in order to get the CPTs. If the calculation can be carried out exactly, they will get the exact form of the importance function as in Equation 9. However, variable elimination is infeasible for large complex networks. Therefore, they set a threshold to the CPT size.

Whenever the size of a CPT generated when eliminating a variable exceeds the threshold, they generate multiple smaller tables to approximate the single big table, so there is no explicit form for their importance function.

For the simple running example, since variable elimination can be carried out exactly, Hernandez et al.'s method is able to generate the exact CPTs.

**Example:** Variable Elimination-based Importance Function for the running example. If we eliminate $B$ before $A$, we get conditional forms $P(A)$ and $P(B|A)$, which are

| $a$ | 0.1033 |
|---|---|
| $\neg a$ | 0.8967 |

| $P(B|A)$ | $a$ | $\neg a$ |
|---|---|---|
| $b$ | 0.0232 | 0.9545 |
| $\neg b$ | 0.9768 | 0.0455 |

□

A simple calculation shows that the importance function is indeed equivalent to the actual posterior distribution. However, for larger models, variable elimination-based importance function often need to use several tables to approximate a single big table. Since the approximation is driven mostly by table size, the approximation can be also sub-optimal. Salmeron et al. [17] later propose to improve the approximation using probability trees to represent the CPTs.

## 5 An Influence-Based Approximation Strategy for Importance Functions

In the previous section, we argued that some approximations may not be able to approximate the posterior distribution well. We now begin to discuss one approximation strategy that is based on the influence among variables in a Bayesian network.

First, we provide an analysis of the influence of evidence. We know that diagnostic evidence makes the ancestors of evidence nodes conditionally dependent. We need to model the most important dependence relations in order to obtain a good importance function. One useful measure to model the relative strength of the dependence relations among the variables is the *sensitivity range* of the probability of an event $y$ with respect to an event $x$ [7]. More formally, suppose that $E = e$ is the observed evidence which might affect the assessment of the probability of $x$, giving $P(x|e)$. Suppose that $Y$ is conditionally independent of $E$ given $X$. Then the sensitivity range is defined as the derivative of $P(y|e)$ with respect to $P(x|e)$:

$$SR(y, x) \equiv \frac{\partial P(y|e)}{\partial P(x|e)} . \quad (12)$$

Henrion [7] has shown that for the causal link in Figure 2, the sensitivity range $SR(y, x)$ with respect to $e$ satisfies the following inequality.

$$|SR(y, x)| \leq 1. \quad (13)$$

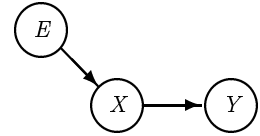

Figure 2: A causal link.

Essentially, Henrion shows that the evidence on a node has more influence on its immediate children than its further descendants. Now, we extend the result to more general scenarios. First, let us look at the diagnostic link in Figure 3. Given conditional independence, $P(y|x) = P(y|x, e)$. We have

$$P(y|e) = P(y|x)P(x|e) + P(y|\neg x)(1 - P(x|e)) . \quad (14)$$

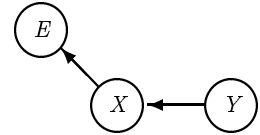

Figure 3: A diagnostic link.

Taking the derivative with respect to $P(x|e)$, we get,

$$SR(y, x) = P(y|x) - P(y|\neg x) . \quad (15)$$

Obviously, Equation 13 also holds for the diagnostic link. It shows that the evidence on a node has more influence on its immediate parents than its further ancestors. Now, let us look at a more interesting case. Suppose we have an intercausal network as in Figure 4, we have

$$\begin{aligned} P(y|z, e) &= P(y|x)P(x|z, e) + \\ &\quad P(y|\neg x)(1 - P(x|z, e)) . \end{aligned} \quad (16)$$

Taking the derivative with respect to $P(x|z, e)$, we get,

$$SR(y, x) = P(y|x) - P(y|\neg x) . \quad (17)$$

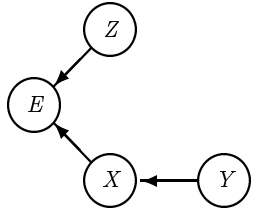

Figure 4: An intercausal link.

Again, Equation 13 also holds for this case. This result shows that, although an evidence node introduces dependence among its ancestors, the strength of the dependence will become weaker as the distance between the ancestors increases.

To summarize, our discussion essentially shows that, in general, as the distance from a variable to the evidence becomes larger in a Bayesian network, the evidence usually has less influence on the posterior distribution of the variable. Also, the dependence relations among the immediate parents of an evidence node are stronger than those among its further ancestors. We can also show the same results using another dependence measure called *mutual information*, which we omit for brevity.

For each CPT $P(X_i|PA(X_i), \mathbf{E}, RF(X_i))$ in Equation 9, $PA(X_i)$ are immediate parents of $X_i$, so their influence are usually strong. $E$ contains observed variables, so it only reduces the complexity of the CPT. However, the variables in $RF(X_i)$ have varying distances from $X_i$. From our analysis, we believe that by throwing away the variables that are further away from $X_i$, we can still reserve a good approximation of the original CPT. Therefore, we propose to approximate the full importance function by adding additional arcs only among the parents of evidence. By modeling the most important additional dependence among the variables, we can anticipate that the importance function can be quite close to the actual posterior distribution. Keeping adding arcs makes the network more complex, which may only bring minimal improvement but only makes the computation more costly.

## 6 Experimental Results

To justify our proposed approximation strategy, we tested it on the EPIS-BN algorithm. We performed our experiments on the ANDES [2], CPCS [15], and PATHFINDER [5] networks. Our comparison was based on the average *Hellinger's distance* [9] between exact posterior marginals of all unobserved variables and sampling results. Hellinger's distance yields identical results as Kullback-Leibler divergence in most cases, but its major advantage is that it can handle zero probabilities, which are common in Bayesian networks. We implemented our algorithm in C++ and performed our tests on a 2.8 GHz Xeon Windows XP computer with 2GB memory.

### 6.1 Review of the Performance of EPIS-BN

The EPIS-BN algorithm is proposed by Yuan and Druzdzel in [20], whose main idea is to use several steps of *loopy belief propagation* (LBP) [10] to estimate an importance function for importance sampling. Experimental results in [20] show that the EPIS-BN algorithm improves upon LBP and achieves a considerable improvement over the state of the art algorithm then, the AIS-BN [1]. Furthermore, the results also show that the EPIS-BN algorithm already approaches the limit that sampling algorithms can achieve on CPCS and PATHFINDER, because the precision that it achieves on these networks is already in the same order as those of probabilistic logic sampling on the same networks without evidence. In the latter case, since there is no evidence in the networks, logic sampling samples from the optimal importance function, the prior distribution. We believe that precision so achieved is the limit of sampling algorithms. However, due to the potential instability of LBP and, hence, possibly poor importance functions, EPIS-BN can still perform sub-optimally, for instance on ANDES.

### 6.2 Results of Different Approximation Strategies on ANDES

In this experiment, we generated a total of 75 test cases for the ANDES network. These cases consisted of five sequences of 15 cases each. For each sequence, we randomly chose a different number of evidence nodes: 15, 20, 25, 30, 35 respectively. We used three different forms of importance function in this experiments. The first one was represented by ICPTs, as in Equation 11. For the second one, we only added additional arcs between the parents of the evidence nodes. For the third one, we carried out Algorithm 1 fully and added all the necessary arcs. We then ran EPIS-BN on the three importance functions. The results are shown in Figure 5.

As expected, different representations of the importance functions yielded different errors. Adding arcs between the parents of evidence nodes brings considerable reduction in error. A paired one-tail t-test at $p = 0.00029$ level shows that the improvement is significant. Since the new importance function has few new arcs, its influence on the running time was minimal. Adding more arcs to get the exact importance function form did not improve the results, but only made the algorithm less efficient. P-value in this case is 0.0018. The results clearly agree with our analysis of the influence of evidence.

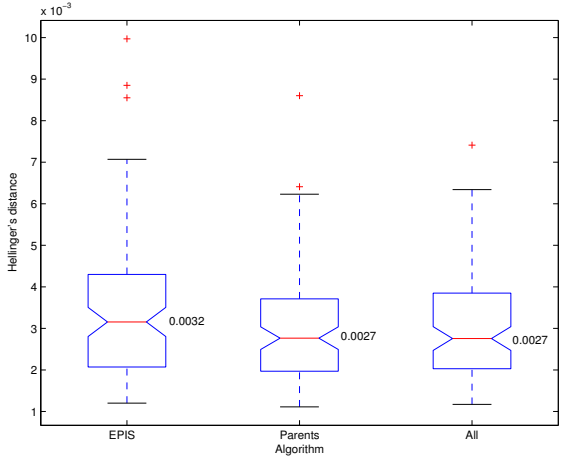

Figure 5: Hellinger's distance of the EPIS-BN algorithm on three different forms of importance function on ANDES. *Parents* stands for the importance function with additional arcs between parents of evidence. *All* stands for the importance function with all additional arcs. Numbers beside the boxplots are the median errors.

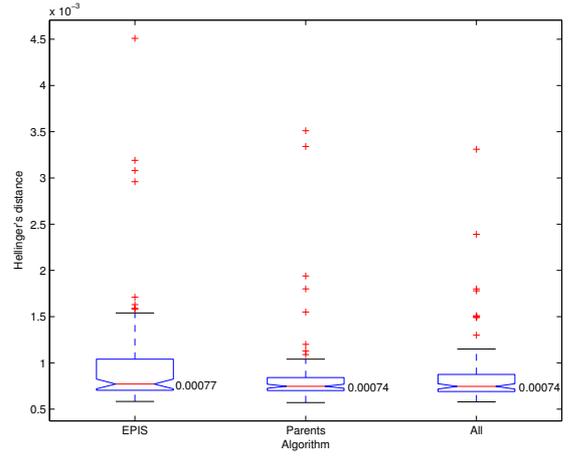

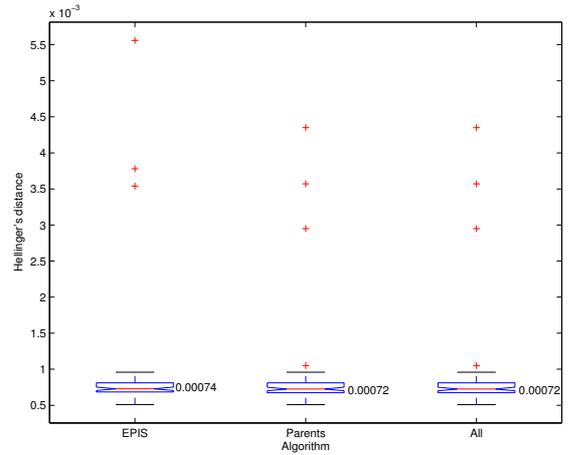

Figure 6: Hellinger's distance of the EPIS-BN algorithm on three different forms of importance function on (**a**) CPCS and (**b**) PathFinder.

### 6.3 Results of Different Approximation Strategies on CPCS and PathFinder

In this experiment, we used the same experimental setup and did some experiments on the CPCS and PathFinder networks. The results are shown in Figure 6. Again, adding arcs among the parents of evidence nodes brings immediate improvements for EPIS-BN. Adding more arcs to get the exact importance function form did not improve the results, but only made the algorithm less efficient.

## 7 Conclusion

In this paper, we address a key problem of importance sampling in Bayesian networks, the representation of the importance function. Typically, we represent an importance function as a factorization, i.e., a sequence of conditional probability tables (CPTs). We usually cannot afford to calculate and store the exact forms of the CPTs. Therefore, different approximations have been taken. We reviewed several popular approximation strategies for the CPTs and point out their limitations. After that, based on an analysis of the influence of evidence in Bayesian networks, we propose an approximation strategy that aims at accommodating the most important additional dependence introduced by the evidence. The proposed importance function is easier to interpret. Our experimental results also show that the new approximation strategy offers an immediate improvement of the quality of the importance function. By introducing more parameters, the improved importance function form also brings much potential for dynamic sampling algorithms, as they can learn theoretically better importance functions.

### Acknowledgements

This research was supported by the Air Force Office of Scientific Research grants F49620–03–1–0187. We thank several anonymous reviewers of the UAI05 conference for several insightful comments that led to improvements in the paper. All experimental data have been obtained using SMILE, a Bayesian inference en-gine developed at the Decision Systems Laboratory and available at http://www.sis.pitt.edu/~genie.